\documentclass[
]{ceurart}

\usepackage{wrapfig}

\usepackage{caption}
\begin{document}

\copyrightyear{2021}
\copyrightclause{Copyright for this paper by its authors.
  Use permitted under Creative Commons License Attribution 4.0
  International (CC BY 4.0).}

\conference{Proceedings of the Text2Story'22 Workshop, Stavanger (Norway), 10-April-2022}

\title{Understanding COVID-19 News Coverage using Medical NLP}

\author[1]{Ali Emre Varol}[%
orcid=0000-0003-3228-1129,
email=emre@johnsnowlabs.com
]
  
\author[1]{Veysel Kocaman}[%
orcid=0000-0002-0065-6478,
email=veysel@johnsnowlabs.com
]

\author[1]{Hasham Ul Haq}[%
orcid=0000-0002-8417-3288,
email=hasham@johnsnowlabs.com
]

\author[1]{David Talby}[%
orcid=0000-0003-2782-5478,
email=david@johnsnowlabs.com
]
\address[1]{John Snow Labs inc. 16192 Coastal Highway,
Lewes, DE 19958, USA}

\begin{abstract}
  Being a global pandemic, the COVID-19 outbreak received global media attention. In this study, we analyze news publications from CNN and The Guardian - two of the world’s most influential media organizations. The dataset includes more than 36,000 articles, analyzed using the clinical and biomedical Natural Language Processing (NLP) models from the Spark NLP for Healthcare library, which enables a deeper analysis of medical concepts than previously achieved. The analysis covers key entities and phrases, observed biases, and change over time in news coverage by correlating mined medical symptoms, procedures, drugs, and guidance with commonly mentioned demographic and occupational groups. Another analysis is of extracted Adverse Drug Events about drug and vaccine manufacturers, which when reported by major news outlets has an impact on vaccine hesitancy.
\end{abstract}


\maketitle

\section{Introduction}
The novel severe acute respiratory syndrome coronavirus 2 (SARS-CoV-2) - commonly known as COVID-19 - was first reported in December 2019. Due to its highly contagious nature, it quickly spread through the world, prompting the World Health Organization to declare the virus outbreak as a global pandemic on March 11, 2020 \cite{world2020corona}. 
As news media reporting is understood to play a central role during national security and health emergencies \cite{laing2012h1n1}, during the pandemic, media representations of complex, rapidly evolving epidemiological science shape public understandings of the risks, measures to limit disease spread, and associated political and policy discourses \cite{mach2021news}. 


In this study we analyse news coverage from two prominent media outlets: CNN and The Guardian. The dataset includes more than 36,000 articles and is analyzed using the clinical and biomedical NLP models from the Spark NLP for Healthcare library  \cite{KOCAMAN2021100058}. Spark NLP is an open-source and widely deployed software library, built on top of Apache Spark, that provides production-grade implementations of recent deep learning and transfer learning NLP algorithms and models. It enables combining  tasks into unified NLP pipelines in Python, R, Java, or Scala and is the only library that can scale up training and inference on any Spark cluster. Spark NLP for Healthcare provides healthcare-specific algorithms and over 400 pre-trained models that have obtained state-of-the-art accuracy on public academic benchmarks in biomedical named entity recognition \cite{kocaman2021biomedical}, clinical assertion status detection \cite{kocaman2020improving}, medical relation extraction   \cite{haq2021deeper}, and adverse event detection \cite{haq2022mining}.

COVID-19 has been studied extensively - including the impact of its news coverage. In previous studies, \cite{info:doi/10.2196/28253} and \cite{9366469} used topic modelling and sentiment analysis, while \cite{politiscovid21} analyzed discourse structures to study bias in media reporting to influence government policies and global reporting. Cresswell et al. \cite{info:doi/10.2196/26618} used social media data to analyse public sentiment towards the pandemic and determined media reporting played a major role in public sentiment. This study is unique in applying healthcare-specific deep learning networks, models, and embeddings - enabling the extraction and correlation of over 100 different types of medical entities with high accuracy. Following are the contributions of this study:
\begin{itemize}
  \item Using fine-tuned NLP models to find most prevalent covid symptoms, prevention guidelines, research institutions, covid variants, and other entities for analysis.
  \item Geographical and demographic analysis of news coverage to understand most affected countries, population age groups, and professions.
  \item Putting the entire covid coverage of 2020 \& 2021 on a timeline for each country, to understand temporal variation and correlation of case count and media coverage.
  \item Comparing the findings from unstructured text with the statistical data reported by WHO and analysing coherency. 
  \item Correlating adverse reactions and drug brands by extracting and linking drug and reaction entities.
\end{itemize}

\section{Analysis}

The dataset comprises of 36,354 live blogs and key moments about COVID-19, published in 2020 and 2021. A live blog is a web page where news media outlets offer daily live coverage about an ongoing event. Each live blog consists of news stories and key moments. Journalists manually select the key moment stories from the whole set of news articles. We excluded the key moments from live blogs; hence, we can appropriately compare live blogs with key moments; otherwise, there will be some overlapping parts. To scrape the news articles, we leveraged web scraping algorithms by \cite{pasquali2021tls}, and updated according to our use-case. Updated code and Spark NLP pipelines can be accessed \cite{code}. As stated in \cite{pasquali2021tls}, first, we collected the live blogs and then applied some pre-processing steps such as parsing HTML format, converting the dates, and extracting key moments from live blogs. The data is organized by title, text, date, URL, and key moment status fields. After data preparation, the executed NLP pipeline included sentence segmentation, tokenization, calculating embeddings, multiple named entity recognition steps, and relation extraction.

In the following sections, we'll report the analysis outcomes that we run over the news coverage by age demographics, news cycle evolution over time and the reporting and impact of adverse drug \& vaccine events.

\subsection{General News Analysis}

\begin{figure}[t!]
    \centering
    \includegraphics[width=\textwidth]{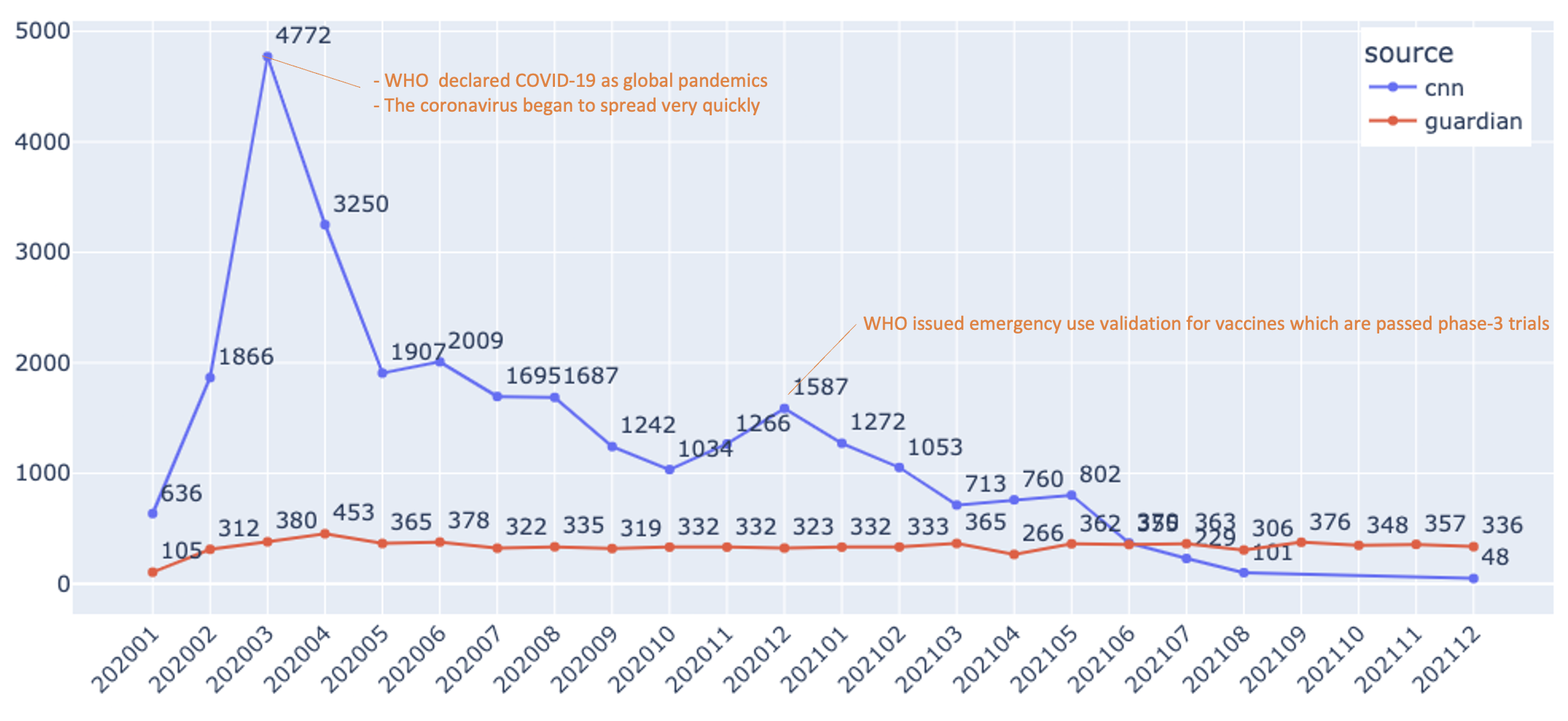}
    \caption{Monthly Distribution of the News from Jan'20 - Dec'21.}
    \label{fig:fig3}
\end{figure}

We tracked the live blogs of CNN from 22nd January 2020 and The Guardian from 24th January 2020; until the end of 2021. The distribution of news by sites and years is explained in Figure \ref{fig:fig3}. Upon inspection, it is apparent that the number of news was high in the periods when COVID-19 first appeared and the first deaths had been reported in and outside Asia in early 2020. News of CNN reached its peak point in March 2020 when COVID-19 reached the shores of United States. Other factors of high news coverage included the massive spread in Europe (especially Italy) and promulgation of the virus as a pandemic by the WHO. Conspicuously, CNN COVID-19 reporting gradually decreased after their peak point and became very seldom in late 2021. With all these, as an overall evaluation, a monthly average of COVID-19 live blogs and key moments is 807, and the median is 365.


If we look at the types of news, it primarily consists of live blogs while the key moments constitute for a minority portion. When we combine them for analysis, we get more than 5.3M words in the corpus to create a story.

\subsection{Demographics and Geographical Analysis by Named Entities}

\begin{figure}[t!]
    \centering
    \includegraphics[width=\textwidth, height=10cm,keepaspectratio]{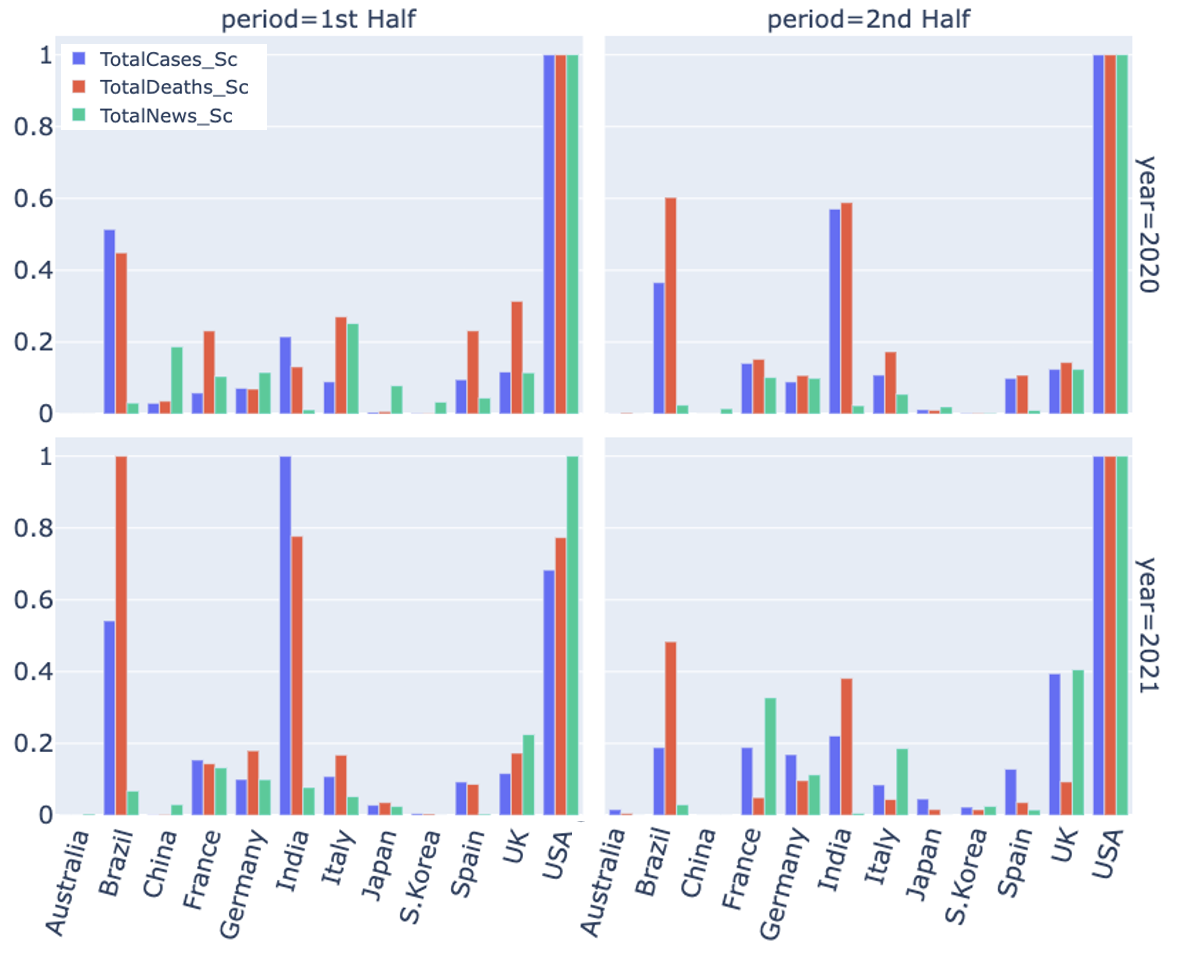}
    \caption{Total number of Cases, Deaths, and News (Scaled by Periods)}
    \label{fig:fig5}
\end{figure}


In order to analyse the news coverage by certain geographical entities, we used a pre-trained named entity recognition (NER) model called \textit{ner\_deid}\footnote{\url{https://nlp.johnsnowlabs.com/2021/09/03/ner_deid_subentity_augmented_en.html}} from Spark NLP for Healthcare library to extract country names mentioned in the news. Using WHO’s official daily statistics \cite{whodataset}, we gathered the number of actual COVID-19 cases and deaths for every half of 2020 and 2021, and then compared with the top 12 countries extracted from the news by the \textit{ner\_deid} model. 

Figure \ref{fig:fig5} shows the normalized metric for each country by dividing its value by the maximum value in the respective six-month period.  As it is seen on the chart, the most reported cases are seen in the USA, India, and Brazil, respectively. It is also seen that the number of patients who died is also among these three countries, although Brazil ranks second on the death list. 


The graph can be interpreted in the following way: (1) If the lengths of the bars are in agreement with each other, it can be deduced that the cases reported and deaths reported are in agreement with the number of news broadcasts in CNN and the Guardian. (2) If the blue and red bars are compatible with each other but not with the green bar, there may be two reasons for a mismatch with the news number. The first reason could be that the country is far from the news source while the second reason could be that the total number of cases and deaths in the country are not considered remarkable by the news sources. (3) If the all the bars are inconsistent with each other, it may indicate that the country in question has not published the case and death numbers transparently. (4) Needless to say, the numbers regarding the countries having inconsistent figures (e.g. Brazil, Russia etc.) might also be explained by the fact that the dataset only covers the news articles from the US (CNN) and UK (Guardian), hence these countries are reported through the lens of other countries (US and UK editors).

Consequently, we can count United States, United Kingdom, France, Germany and Australia as the most compatible countries while China and Brazil can be counted as the most incompatible countries. For Spain, the total number of cases and deaths is consistent, but the volume of news articles is inconsistent. This may be explained by Spain being away from the two news sources.

 \begin{figure}
  \centering
  \includegraphics[width=\linewidth]{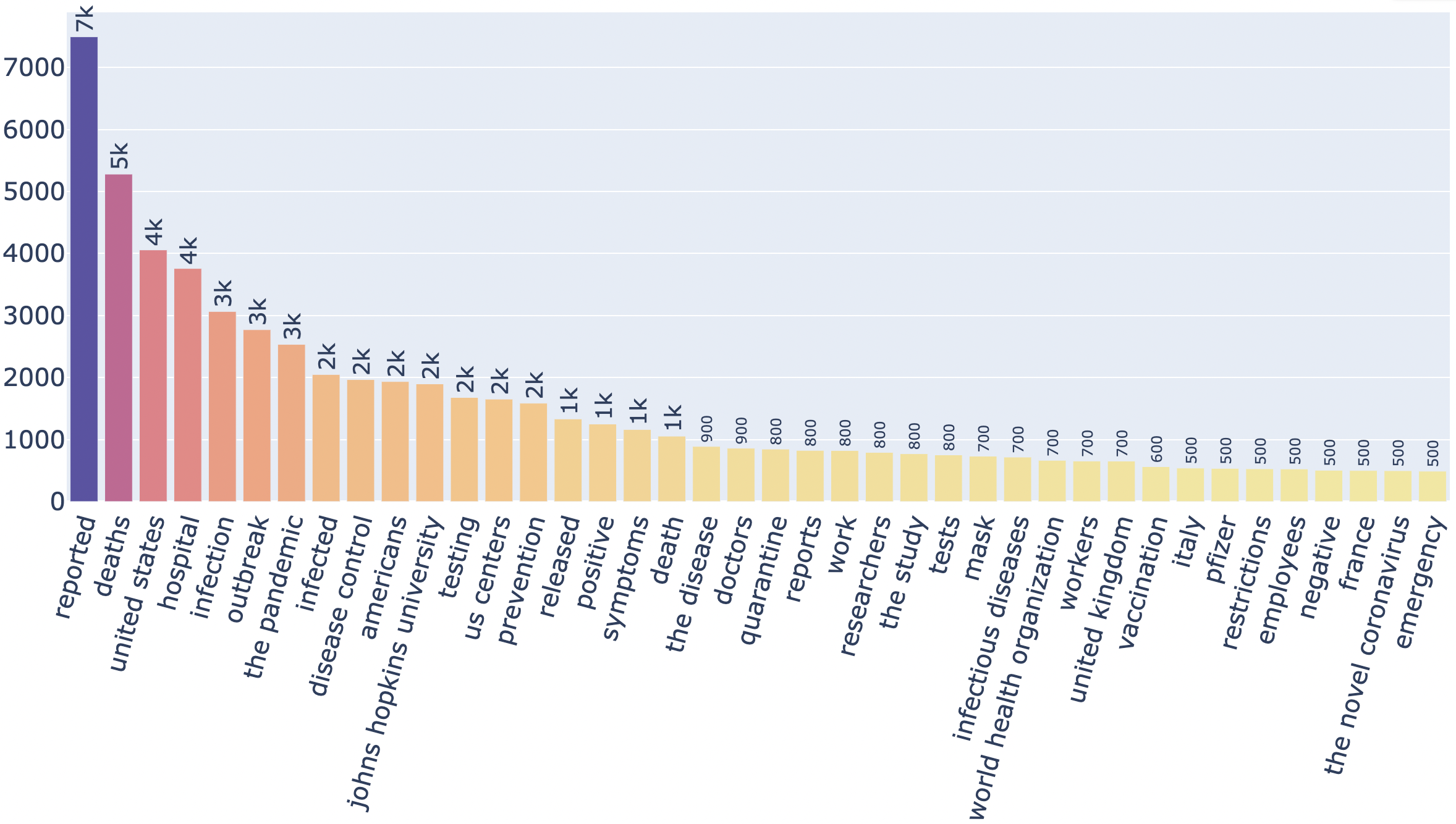}
  \caption{Top 40 Entities in Corpus}
  \label{fig:fig8}
 \end{figure}
 
\subsection{Behavioral and Clinical Analysis by  Named Entities}

In order to analyse the news with respect to the clinical entities mentioned, we used a pretrained NER model named \textit{ner\_events\_clinical}\footnote{\url{https://nlp.johnsnowlabs.com/2021/03/31/ner_events_clinical_en.html}} and  \textit{ner\_jsl}\footnote{\url{https://nlp.johnsnowlabs.com/2020/04/22/ner_jsl_en.html}} to extract major clinical events. These two NER models combined can extract more than one hundred different types of clinical entities at once, with one line of code, enabling a deeper analysis than previous studies of this type.

Figure \ref{fig:fig8} shows the top 40 entities by frequencies in the corpus. Entities related to coronavirus (deaths, infections, preventive measures, and geographical locations etc) seem to be the most prevalent, followed by entities related to research and development of drugs. When we look at the ranking of countries in this list, the United States holds first place with the most reported number of cases, while the United Kingdom and Italy are in the second and third places - aligning with the WHO stats illustrated in Figure \ref{fig:fig5}. The only university listed is John Hopkins University (JHU), primarily because of its role in organization and coordination of the Coronavirus Resource Center \cite{jhu_coronavirus}. We also see major vaccine manufacturers in the list due to high coverage of vaccine development and trials.

\begin{figure}
  \centering
  \includegraphics[width=\linewidth]{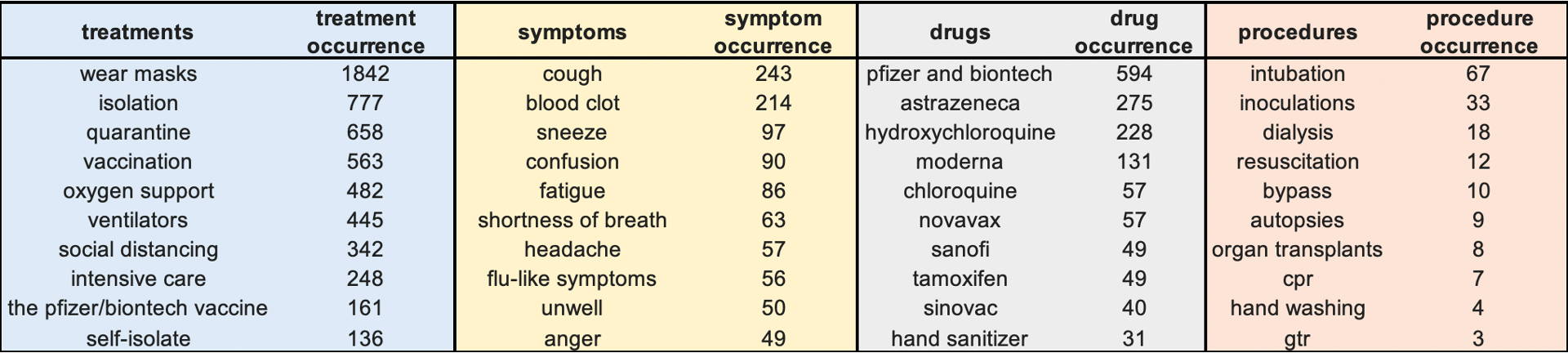}
  \caption{Top 10 Treatments, Symptoms, Drugs, and Procedures}
  \label{fig:fig9_10}
\end{figure}

Figure \ref{fig:fig9_10} shows the Top 10 treatments, symptoms, drugs, and procedures in the order of their frequencies. It is clear that concepts like vaccine, wearing masks, oxygen support, and ventilators appear often in treatments. These entities along with isolation, quarantine, social distancing, and self-isolation, which are at different ranks in the list, have a significant place in disease control. 

On the other hand, the symptoms presented in Figure \ref{fig:fig9_10} are widespread. People who observe any of these symptoms turn to hospitals or health institutions to find out if they have COVID-19. Coughing, blood clotting, sneezing, fatigue, shortness of breath, and headache are the prevailing COVID-19 symptoms. It is worth noting that even a headache alone was enough to arouse the suspicion of coronavirus during this period. However, very specific symptoms related to loss of smell and taste are also observed, although their frequency is lower.

When analyzing named entities related to drugs, the companies producing vaccines stand out. Note that drug names also include drug brand names, drug manufacturers, and drug ingredients. The data suggests that the manufacturer's name is more prominent compared to the drug's scientific name in news coverage. Hydroxychloroquine and chloroquine were granted an emergency use authorization (EUA) on March 28, 2020 for treating COVID-19 cases. However, on June 15, 2020, FDA revoked its use \cite{fda}, limiting it only to hospitalized patients under heart monitoring.

The most commonly mentioned medical procedures feature intubation and inoculation in the top two places. Intubation has been the most frequently applied procedure for patients admitted to the intensive care unit during this period. 

\begin{wrapfigure}{r}{0.40\textwidth}
  \centering
  \includegraphics[width=1.0\linewidth]{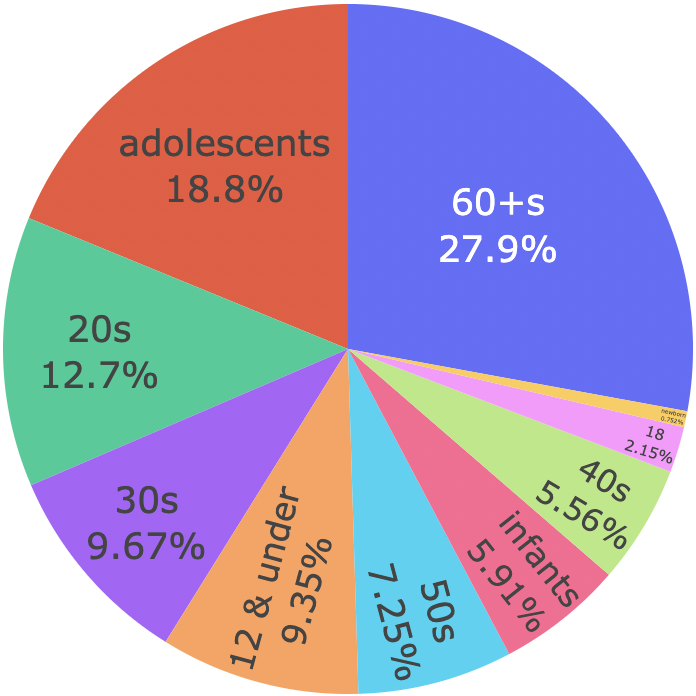}
  \caption{Age groups mentioned in news}
  \label{fig:fig12}
\end{wrapfigure}

The pie chart in Figure \ref{fig:fig12} shows that the most mentioned age group is 60 years and above, followed by adolescents. This is probably due to the fact that COVID-19 has higher severity in older people. According to scientific observations, WHO declared that COVID-19 is often more severe in people who are older than 60 years or who have health conditions like lung or heart disease, diabetes, or conditions that affect their immune system \cite{who_risk_groups}.





Comparing the most prevalent entities in 2020 versus 2021 - the coronavirus, which was in the first place in 2020, fell to second place in 2021, ceding the first place to the COVID-19 vaccine. The word 'reported' fell from the second most encountered entity in 2020 to the fourth in 2021. This decline can be attributed to less willingness to reporting of COVID-19 cases since according the \cite{whocovid} the trend of reported cases is not declining. On the contrary, there is an increase in both the number of new cases and the number of deaths.




In addition to healthcare-specific named entity recognition and resolution models, Spark NLP also provides unsupervised key phrase extraction from free text. 
Results indicate that the entity counts follow the general trend where COVID-19 infection is the most prevalent term, followed by entities related to death, geographical locations, media outlets, vaccine development, and vaccination reports. 

COVID-19 also highlighted specific professions namely the medical doctors and nurses. Our NER analysis show that researchers, workers, teachers and government officials are the most cited professions in the news. Understandably, the most commonly mentioned people in the news during this period are healthcare professionals. 

\subsection{Impact of Adverse Drug and Vaccine Events}

Adverse drug reactions/events (ADR/ADE) have a major impact on patient outcomes and healthcare costs. An analysis of extracted Adverse Drug Events versus drug and vaccine manufacturers reported by major news outlets has an impact on vaccine hesitancy as well as people's reactions to medications and public health measurements dictated by regulatory agencies. In order to analyse ADEs as well the adverse events of vaccines mentioned in the news, we used a pre-trained NLP pipeline named \textit{explain\_clinical\_doc\_ade}\footnote{\url{https://nlp.johnsnowlabs.com/2021/07/15/explain_clinical_doc_ade_en.html}} that comes with the Spark NLP for Healthcare library. This NLP pipeline is the first scalable end-to-end solution for mining ADE’s from unstructured text, including Document Classification, Named Entity Recognition, and Relation Extraction Models within a unified NLP pipeline \cite{haq2022mining}.

\begin{figure}
    \centering
    \begin{minipage}[10cm]{0.50\textwidth}
        \centering
        \includegraphics[width=\textwidth]{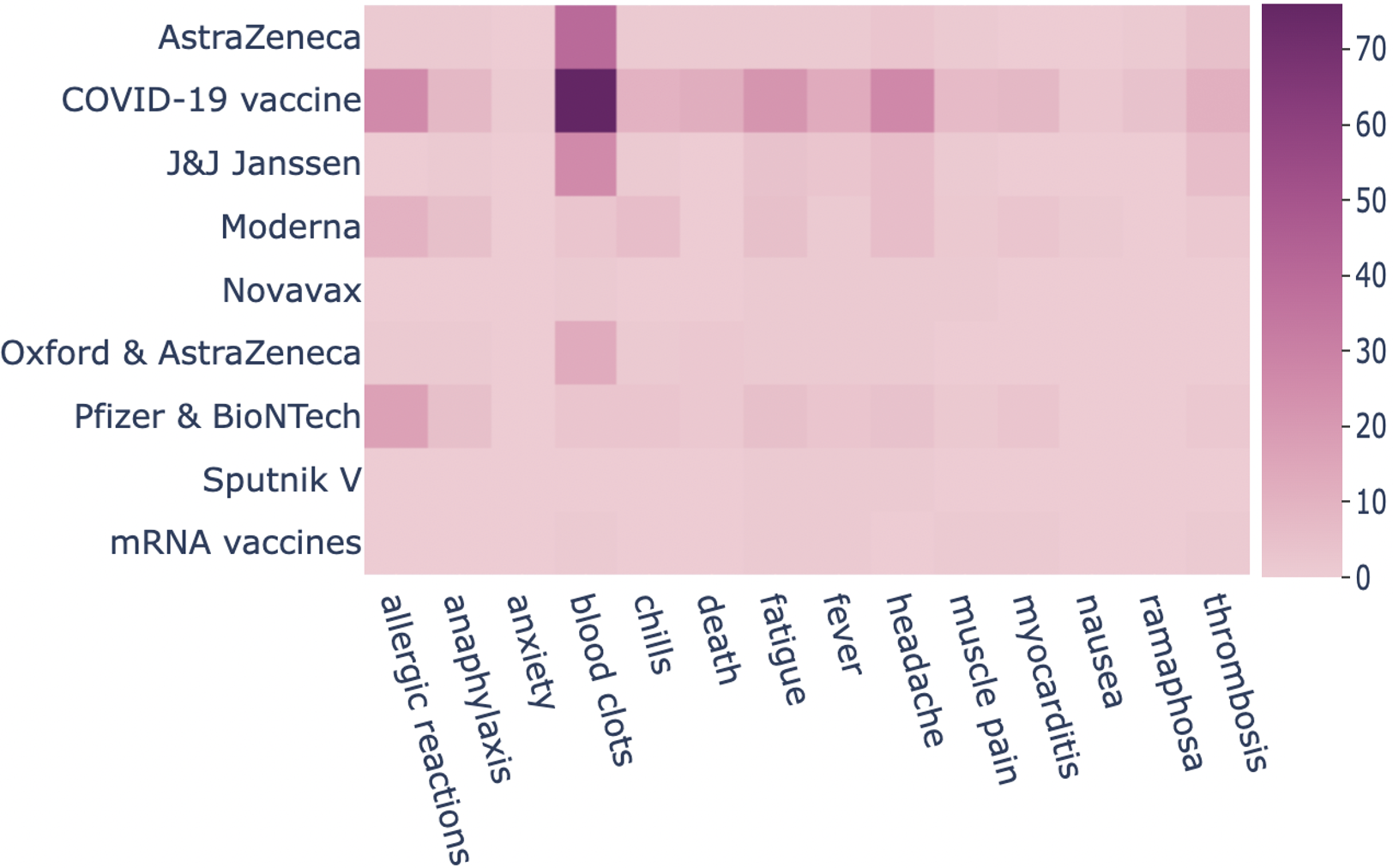} 
        \caption{ADE of Vaccines}
        \label{fig:fig18}
    \end{minipage}\hfill
    \begin{minipage}[10cm]{0.50\textwidth}
        \centering
        \includegraphics[width=\textwidth]{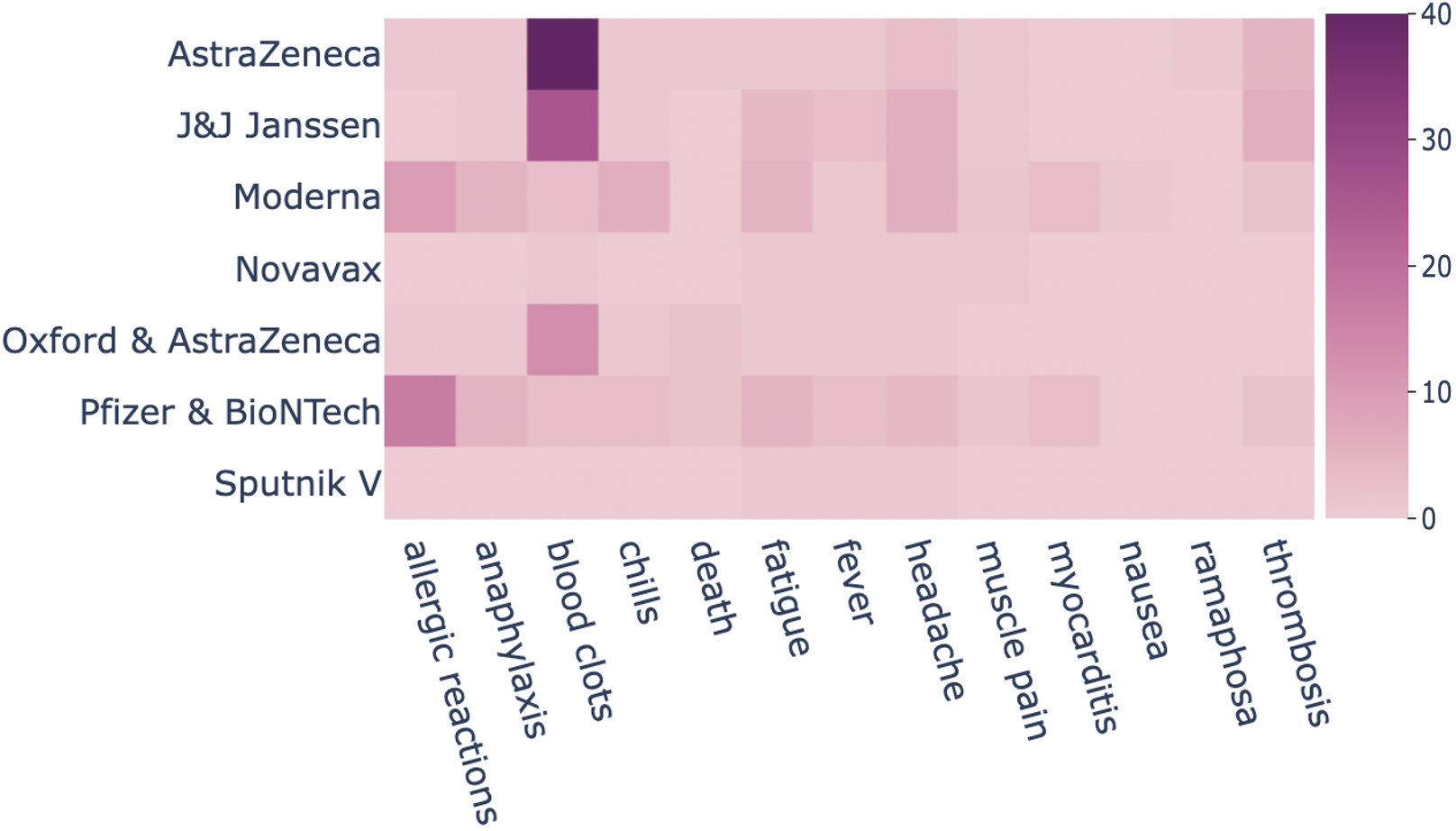}
        \caption{ADE of Vaccines without Generic Titles} 
        \label{fig:fig19}
    \end{minipage}
\end{figure}

Figure \ref{fig:fig18} is a heatmap of correlation between vaccines and adverse reactions. When we explore the news dataset, we see that some of the vaccines are listed as their manufacturer's name, but some are only with generic titles such as COVID-19 vaccine or mRNA vaccines. Since the heavy use of such general entities dilutes the results, Figure \ref{fig:fig19} shows the correlation while eliminating general vaccine names.

One of the most prominent observation the entanglement of clotting with AstraZeneca and Johnson \& Johnson. This phenomenon was widely reported by the media, suspending the use of the drugs by these companies \cite{wise2021covid}. General allergic reactions are more common for vaccines from Pfizer \& BioNTech and Moderna \cite{meo2021covid}.

\section{Conclusion}

As a global pandemic, COVID-19 has caused millions of deaths and impacted most people on earth. The media has played a major role in shaping public awareness, education, and opinion. Due to the growing number of media channels, the constant barrage of news, the prevalence of social media in sharing and shaping the news, and frequent shifts in regulations dictated by governments to combat the pandemic, analysing such a sheer volume of information objectively in a short amount of time can be one of the most practical current applications of NLP. In this study, we analyse news coverage from two prominent media outlets by using the clinical and biomedical NLP models of Spark NLP for Healthcare.

The analysis shows that the total number of cases and deaths reported by the news are mostly consistent with the numbers shared by WHO officially. The pre-trained NLP models perform well on extracting the most relevant terms and entities that have been widely used in the news. One tangible outcome of this study can be the automated mining of the adverse reactions to drugs and vaccines that are used to combat the virus. The analysis shows the correlations between the prominent vaccine manufacturers and reported adverse events just by relying on the news articles used for this study. Future research can broaden this automated analysis to include additional news sources, news in languages other than English, dealing with fake news, or to correlating changes in news coverage to resulting changes in public opinion and behavior.


\bibliography{sample-ceur}




\end{document}